# ISLR101: an Iranian Word-Level Sign Language Recognition Dataset

**Hossein Ranjbar, Alireza Taheri**[*]

Social and Cognitive Robotics Lab., Department of Mechanical Engineering, Sharif University of Technology, Tehran, Iran

{hosein.ranjbar, artaheri}@sharif.edu

*Corresponding Author; Tel: +98 21 6616 5531

**Abstract**

Sign language recognition involves modeling complex multichannel information, such as hand shapes and movements while relying on sufficient sign language-specific data. However, sign languages are often under-resourced, posing a significant challenge for research and development in this field. To address this gap, we introduce ISLR101, the first publicly available[1] Iranian Sign Language dataset for isolated sign language recognition. This comprehensive dataset includes 4,614 videos covering 101 distinct signs, recorded from 10 different signers (3 deaf individuals, 2 sign language interpreters, and 5 L2 learners[2]) against varied backgrounds, with a resolution of 800×600 pixels and a frame rate of 25 frames per second. It also includes skeleton pose information extracted using OpenPose. We establish both a visual appearance-based and a skeleton-based framework as baseline models, thoroughly training and evaluating them on ISLR101. These models achieve 97.01% and 94.02% accuracy on the test set, respectively. Additionally, we publish the train, validation, and test splits to facilitate fair comparisons.

## 1. Introduction

Sign language is the primary means of communication for millions of individuals with hearing impairments worldwide, and sign language technology works to bridge the communication gap between them and hearing people. Sign languages are complex, with large vocabularies and unique phonological rules. Analogous to the sounds of speech, signs are composed of largely discrete elements (e.g., handshape and/or lip shape, location, and movement) according to complex rules [1]. Given the limited familiarity of most hearing individuals with sign language, there is a growing need for Sign Language Recognition (SLR) and Translation (SLT) technologies to facilitate understanding and communication. In the literature, SLR research is divided into two branches. The first is isolated SLR, where a single sign is performed by the signer, and the system maps the spatiotemporal sequence to a specific sign, treating it as a classification problem. The second is continuous SLR, where multiple signs are present in the input sequence, and the system converts the sequence of frames into a sequence of signs, framing it as a sequence-to-sequence problem. In SLT, the given spatiotemporal sequence is mapped and translated into spoken sentences.

---

[1] The dataset will be made available upon request for academic and research purposes.
[2] L2 learners are individuals acquiring sign language as a second language with varying fluency.

Serving as a fundamental building block for understanding sign language sentences, the word-level sign recognition task itself is also very challenging:

- Given the relatively small number of sign language users compared to the general population, it is more challenging to find expert subjects for recording signs.
- Most sign languages lack formal standardization, even within specific regions. Each region often has its own distinct sign language, contributing to the diversity and complexity of sign languages worldwide. A single word in spoken language can have multiple representations in sign language. For instance, in Iranian Sign Language (ISL), the sign "snow" can be signed using either one hand or two hands, as shown in Figure 1a.
- The meaning of signs is largely determined by the interplay of body movements, manual gestures, and facial expressions, where subtle variations can lead to different interpretations. As illustrated in Figure 1b, the signs for "late" and "should" in ISL differ in facial expressions and lip movements.
- Signs are inherently multimodal, necessitating the capture and modeling of various signals to create effective datasets for training SLR systems.

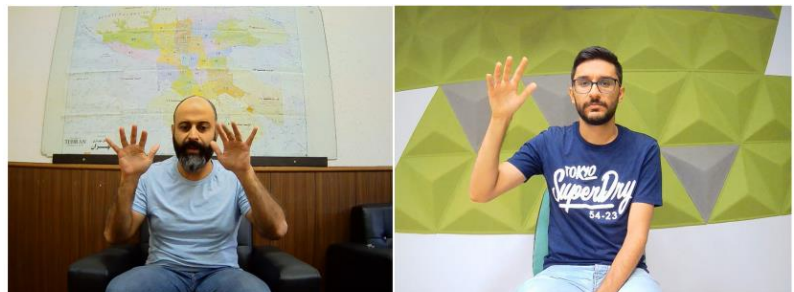
(a) Two Sign Representations of "Snow"

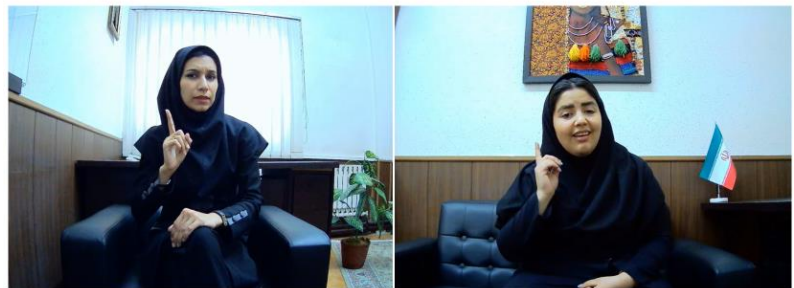
(b) Left: Should, Right: Late

Fig. 1. (a) Two distinct signs for "snow" in ISL, highlighting the lack of formal standardization, and (b) Comparison of the signs for "late" and "should" in ISL, demonstrates how differences in facial expressions and lip movements can alter meaning.

For these reasons, robust SLR systems remain a distant goal. Existing literature has primarily concentrated on extracting multichannel information, and various deep-learning techniques have emerged for sign language recognition. Despite these advancements, the field is still nascent, facing challenges not only in the extraction and modeling of diverse information channels but also due to resource scarcity [2].

In this study, we focus on recording word-level signs in ISL along with their corresponding annotations. To minimize hardware requirements, we utilize only monocular RGB videos, allowing the trained recognition models to function without specialized equipment, such as depth cameras or color gloves. This makes deployment feasible in everyday scenarios. Furthermore, since signers typically communicate in near-frontal views, we ensure that our recordings capture this perspective to enhance dataset quality.

The dataset is called "ISLR101" and contains 4,614 videos, each video only contains one sign in ISL, covering 101 classes, performed by 10 signers. There are more than 40 videos for each word, which is a high ratio compared to other datasets, which makes learning the model well. Each sign is performed by at least 8 different signers. Thus, inter-signer variations in our dataset facilitate the generalization ability of the trained sign recognition models.

Based on ISLR101, we can experiment with several deep learning methods for word-level sign recognition, based on (i) holistic visual appearance, and (ii) 2D human skeleton. For appearance-based methods, we provide a baseline containing three modules: a spatial modeling module that extracts spatial features of each frame, a temporal modeling module that explores relations between frames, and a classification module that assigns a category to the video. We use MobileNet-V2 [3] for the spatial modeling module and Transformer [4] for the temporal modeling module.

For pose-based methods, we first extract human poses from original videos and use them as input features, and we use ST-TR [5] architecture that utilizes spatial self-attention to extract spatial relationships between body key points and temporal self-attention to study the dynamics of each joint along all the frames.

The rest of the paper is organized as follows. Section 2 deals with the related work. Section 3 gives an insight into the datasets. In section 4, we present the baseline models, followed by the experiments and results with discussions in section 5. Finally, we conclude our work in section 5. The primary contributions of this paper are as follows:

- We address the significant challenge of data scarcity in sign language technology by focusing on word-level recognition for Iranian Sign Language (ISL), an area with limited prior research and no publicly available datasets.
- We introduce ISLR101, the first publicly available dataset, to the best of our knowledge, for ISL word-level recognition, comprising 4,614 videos of 101 distinct signs recorded from 10 signers with varied backgrounds, a resolution of 800×600 pixels, a frame rate of 25 frames per second, and accompanying skeleton pose information extracted using OpenPose.
- We establish and benchmark baseline models for ISL recognition using both visual appearance-based and skeleton-based frameworks, achieving high accuracy of 97.01% and 94.02% on the test set, respectively, and release the train, validation, and test splits of ISLR101 to promote transparency and enable fair comparisons in future research.

## 2. Related Work

In this section, we provide a brief overview of existing public sign language datasets and discuss state-of-the-art sign language recognition algorithms. This review highlights the necessity for a large-scale isolated sign language recognition dataset.

### 2.1. Sign Language Datasets

We start by briefly reviewing public benchmarks for automatic sign language recognition. Numerous benchmarks have been proposed for American [6, 7, 8, 9], German [10], Chinese [11, 12], and Polish [13] sign languages. In contrast, datasets for ISL are notably scarce.

Table 1 presents the most prominent video-based, research-level datasets for recognition. Since the dataset described in this paper focuses on sign-level recognition, we only list word-level datasets.

Table 1. Overview of word-level datasets.

| Datasets | Year | Sign Language | Classes | Samples | Signers | Type |
|---|---|---|---|---|---|---|
| RWTH-BOSTON-50 [6] | 2005 | American | 50 | 483 | 3 | RGB |
| Purdue RVL-SLLL [7] | 2006 | American | 39 | 546 | 14 | RGB |
| GSL [14] | 2007 | Greek | 20 | 840 | 6 | RGB |
| Boston ASLLVD [8] | 2008 | American | 2,742 | 9,794 | 6 | RGB |
| DGS Kinect 40 [10] | 2011 | German | 40 | 3,000 | 15 | RGB, depth |
| PSL Kinect 30 [13] | 2014 | Polish | 30 | 300 | 1 | RGB, depth |
| PSL ToF [13] | 2014 | Polish | 84 | 1680 | 1 | RGB, depth |
| DEVISIGN [11] | 2015 | Chinese | 2,000 | 24,000 | 8 | RGB, depth |
| LSE-sign [16] | 2016 | Spanish | 2,400 | 2,400 | 2 | RGB |
| CSL [12] | 2019 | Chinese | 500 | 125,000 | 50 | RGB? |
| WLASL [9] | 2020 | American | 2,000 | 21,083 | 119 | RGB |
| AUTSL [17] | 2020 | Turkish | 226 | 38,336 | 43 | RGB, depth, skeleton |
| RKS-PERSIANSIGN [18] | 2020 | Iranian | 100 | 10,000 | 10 | RGB |
| LSA64 [15] | 2023 | Argentinian | 64 | 3,200 | 10 | RGB |
| **ISLR101 (ours)** | **2024** | **Iranian** | **101** | **4,614** | **10** | **RGB** |

Sign languages vary across regions, each with its unique lexicon and set of signs. As a result, sign language recognition must be approached differently in each context. New movements, handshapes, lip movements, and their combinations necessitate the collection of new training data and may introduce challenges that have not been previously addressed [10, 16]. Consequently, numerous datasets for various sign languages have been compiled and recorded worldwide.

Several datasets for American Sign Language (ASL) have been released, which we review in the following. The RWTH-BOSTON-50 [6] dataset contains 50 signs performed by 3 signers, providing 483 RGB samples in total. Purdue RVL-SLLL [7], published in 2006, includes handshapes, motions, signs, and sentences performed by 14 signers, comprising 2,576 RGB videos. In later years, several large-scale datasets have been published. Table 1 provides an overview of the large-scale isolated sign language datasets.

BOSTON ASLLVD [8] encompasses 2,742 signs, boasting a substantial vocabulary size. However, the dataset contains only 9,794 samples in total, averaging merely 3.6 examples per sign. This dataset was developed to support sign lookup technologies for ASL and includes video sequences captured from four cameras simultaneously: two frontal views, one side view, and one zoomed-in on the signer's face. WLASL [9] is another ASL dataset that offers four different vocabulary sizes, the largest containing 2,000 signs (WLASL-2000 in our tables). It includes 21,083 video samples of 2,000 unique signs, performed by 119 signers. Each sign is performed by at least 3 different signers. The dataset consists of only RGB videos, sourced from 20 educational sign language websites that provide ASL lookup functions, as well as ASL tutorial videos on YouTube. The videos typically feature signers in a near-frontal view, set against plain backgrounds, often wearing black clothing to enhance the visibility of the signs.

PSL Kinect 30 [13] and PSL ToF 84 [13] are Polish Sign Language datasets that consist of 30 and 84 signs, and in total 300 and 1680 samples, respectively. Both datasets provide RGB and depth modalities. LSA [15], an Argentinian Sign Language dataset, includes 64 signs performed by 10 signers, amounting to 3,200 RGB samples. Signers wore different colored gloves for each hand during recording to enhance hand distinction. DEVISIGN [11] is a Chinese Sign Language dataset that consists of 2,000 signs and 24,000 samples that are performed by 8 non-native signers. The videos are recorded in a controlled lab environment with Microsoft Kinect v1, which provides RGB, depth, and skeleton data, in a lab environment in front of a white wall. CSL [12] is a Chinese Sign Language dataset that consists of 500 signs performed by 50 different signers and 125,000 samples. It is recorded with Microsoft Kinect v2 which provides RGB, depth, and skeleton data. Besides being large-scale, this dataset also focuses on user-independent recognition of signs. They select different signers for the training and test sets. The videos are recorded in front of a white background.

To the best of our knowledge, no publicly available datasets exist for Iranian Sign Language (ISL). RKS-PERSIANSIGN [18] is the only video-based ISL recognition dataset, containing 100 words (i.e., glosses) with 10,000 examples (approximately 100 examples per gloss), but it has not yet been released. Additionally, a few dictionaries focus on teaching the language; however, these were recorded for instructional purposes and typically include only one sample per sign, low image quality, and poor annotations, making them unsuitable for training automatic recognition systems. This highlights the need for a research-level dataset that adequately represents the range of signs used in ISL.

Our ISLR101 dataset is a new large-scale Iranian Sign Language dataset comprising 101 signs and 4,614 samples in total, performed by 10 different signers. With an average of 45.7 samples per sign, it provides a substantial number of examples per sign compared to other datasets. Unlike many large-scale datasets, ISLR101 presents varied backgrounds and numerous challenges, such as lighting variations and diverse indoor settings, aiming to reflect realistic, everyday use cases. Detailed information about ISLR101 is provided in Section 3. Additionally, the dataset becomes publicly available, and we also offer skeleton data extracted using OpenPose [20] for researchers working on skeleton-based frameworks.

## 2.2. Isolated Sign Language Recognition Frameworks

Early approaches to sign language recognition relied on hand-crafted features to represent static hand poses, utilizing techniques such as SIFT-based features [21, 22], HOG-based features [23, 24], and frequency domain features [25]. Hidden Markov Models (HMM) [26] were commonly applied to model temporal relationships in video sequences, while Dynamic Time Warping (DTW) [27] was used to manage variations in sequence length and frame rates. Classification methods, like Support Vector Machines (SVM) [28], were then employed to map the extracted features to corresponding signs. However, over the past decade, research has shifted towards end-to-end deep learning, driven by the success of Convolutional Neural Networks (CNNs) for computer vision tasks and Recurrent Neural Networks (RNNs)/Transformers for sequence processing. This transition has led to significant progress in sign language recognition. In the following section, we review deep learning-based frameworks for sign language recognition, categorized into two groups: visual appearance-based and skeleton-based approaches.

### 2.2.1 Visual Appearance-Based Frameworks (VABF)

Most previous works employ an architecture comprising three key modules: a spatial modeling module (e.g., using CNNs [29, 30] or Vision Transformer [31]), a temporal modeling module (utilizing RNNs [29], 1D-CNNs [32], and/or Transformers [31]), and a classification module (typically using fully connected layers or SVMs). Notably, 3D-CNNs [33] have shown particular effectiveness in action recognition tasks due to their ability to capture both spatial and temporal information through 3D convolutions and pooling. These 3D-CNN-based architectures have set the state-of-the-art in several computer vision domains, including action recognition [34, 35] and sign language recognition [12, 36].

The work introducing the WLASL dataset for isolated sign recognition evaluates several deep learning architectures, including a 2D CNN followed by an RNN, a 3D CNN, a pose RNN, and a pose Temporal Graph Convolutional Network [9]. The pose-based networks utilize OpenPose keypoints as input features. Among these, the 3D CNN, specifically the I3D model, achieves the best performance.

After the successful adoption of Transformers in various sequence modeling tasks, including neural machine translation [37] and speech recognition [38], their capabilities to effectively model global contexts make it logical to introduce Transformers to SLR [39, 30] for temporal modeling, as an alternative to traditional RNNs (LSTMs [40] and GRUs) and 1D-CNNs.

De Coster et al. [41] utilize a video transformer architecture [42], which comprises a spatial feature extractor (ResNet-34) followed by a self-attention encoder with four layers. To enhance supervision, they also incorporate hand cropping and pose flows. Additionally, De Coster et al. [43] developed four models to recognize isolated signs in the Flemish Sign Language corpus, using OpenPose and ResNet-34 [44] as feature extractors. In the first model, keypoints extracted from OpenPose are fed into an LSTM. The second model replaces the LSTM with a Transformer for temporal modeling. The third model employs ResNet-34 in the spatial modeling module and a Transformer for temporal modeling. Finally, the fourth model concatenates OpenPose features with spatial features from ResNet-34, also utilizing a Transformer for temporal modeling. Notably, the accuracy improves progressively from the first to the fourth model.

The realm of spatial modeling in computer vision was long dominated by CNNs for an extended period. In recent times, the computer vision landscape has witnessed a significant transformation with the emergence of Vision Transformer (ViT) [45], marking a paradigm shift in spatial modeling. ViT is a pioneering effort to demonstrate how transformers can fully supplant standard convolutions in deep neural networks on large-scale image datasets. They applied the original Transformer model with minimal modifications to a sequence of image 'patches' flattened as vectors. Since transformer architectures do not inherently encode inductive biases (prior knowledge) to deal with visual data, they typically demand a substantial amount of training data to figure out the underlying modality-specific rules. To address this limitation, several attempts have been made to include inductive biases (commonly used in CNNs) alongside the transformers to improve their efficiency [46, 47].

Despite the success of vision transformer models in addressing high-level vision problems, their adoption in the visual modules of SLR has been limited. Kothadiya et al. [48] employed ViT for static Indian sign language recognition. Similarly, Xiao et al. [49] introduced a novel framework for continuous sign language recognition (SLRFormer) based on a Vision Transformer. This framework consists of two key modules: the visual feature extraction module and the temporal information encoding module. In this approach, the ViT serves as the visual feature extractor, while the features extracted by the Vision Transformer are subsequently fed into a bi-directional LSTM for temporal modeling. The output is then enhanced by a weight-attention mechanism that refines the generated results.

Zuo et al. [50] argue that leveraging semantic similarity in English glosses can enhance SLR, as signs that are related in meaning often share phonological characteristics. However, to substantiate this claim, their study primarily relies on internet-scraped datasets, including WLASL, which utilize English glosses to merge and differentiate signs. Their analysis of this dataset reveals that this methodology may introduce artifacts, where distinct glosses correspond to the same signs in ASL. Consequently, it remains unclear whether the improvements observed from their proposed method stem from rectifying these artifacts or from the inherent linguistic properties of sign languages.

### 2.2.2 Skeleton-Based Frameworks

Visual appearance-based models often struggle with significant variations in lighting conditions, background clutter, and occlusions. These models may struggle to generalize to new or unseen actions that significantly differ in appearance from the training data. By abstracting away visual appearance and focusing solely on skeletal information, these models are less sensitive to these variations. Skeleton data, which typically comprises a set of joint coordinates, provides a more compact and efficient representation compared to raw pixel data or video frames. This abstraction can result in faster inference times and reduced computational complexity.

With the emergence of off-the-shelf pre-trained human pose estimation systems like OpenPose, numerous researchers in sign language recognition have employed recurrent neural networks that utilize keypoints as input features [43, 49]. However, because movements in sign language can be quick (leading to motion blur), and because there is occlusion between and within the hands, these keypoints can be noisy [10]. Furthermore, recent works have shown that end-to-end models can significantly outperform pose-based models [9, 51].

The dynamic skeleton modality can be naturally represented by a time series of human joint locations, in the form of 2D or 3D coordinates. Human actions can then be recognized by analyzing the motion patterns thereof. Earlier methods of using skeletons for sign language recognition simply employ the joint coordinates at individual time steps to form feature vectors and apply temporal analysis thereon [30]. However, the effectiveness of these methods is limited because they do not explicitly exploit the spatial relationships among the joints, which are essential for accurately interpreting human actions. Recently, new techniques have emerged that aim to leverage these natural connections between joints for improved performance.

Currently, Graph Neural Networks (GNNs), particularly Graph Convolutional Networks (GCNs), have become the predominant method for skeleton-based SLR. Their ability to efficiently represent non-Euclidean data allows them to effectively capture both spatial (intra-frame) and temporal (inter-frame) information. GCNs were first introduced in skeleton-based action recognition by Yan et al. [44] and are commonly referred to as Spatial-Temporal Graph Convolutional Networks (ST-GCNs). These models process spatial information by leveraging the connections between skeleton joints, while temporal information is captured through additional time connections between each joint across frames.

Despite being proven to perform very well on skeleton data, ST-GCN models have some structural limitations. To address these limitations, Plizzari et al. [53] propose a novel Spatial-Temporal Transformer network (ST-TR), which utilizes the self-attention mechanism of Transformers to model dependencies between joints. In the ST-TR model, a Spatial Self-Attention module (SSA) is used to understand intra-frame interactions between different body parts, and a Temporal Self-Attention module (TSA) to model inter-frame correlations. These two modules are integrated within a two-stream network architecture. In [54], spatial and temporal information has been captured separately using GCNs and a BERT model, and late fusion is performed to make the final predictions.

## 3. ISLR101 Dataset

In this section, we present our Iranian Sign Language dataset, referred to as ISLR101, which comprises 101 distinct signs. Our primary motivation for this collection is to create a comprehensive dataset featuring diverse and challenging backgrounds, conducive to developing a realistic SLR system that reflects real-life scenarios. The dataset was recorded using a K04 2K/4MP streaming webcam, capturing footage at a resolution of 800×600 pixels and a frame rate of 25 frames per second. The camera is positioned approximately 1.5 meters away from the signer to ensure optimal visibility and detail in the recorded gestures. Videos range from a maximum length of 112 frames (4.48 seconds) to a minimum of 5 frames (0.20 seconds), with an average length of 58.02 frames (2.27 seconds).

In selecting the signs for our dataset, we prioritized those that are frequently used in everyday academic discourse. This focus allows for the construction of various sentences and enhances the dataset's applicability for continuous sign language recognition.

The selected signs encompass a broad range of hand/lip shapes and movements. The names of the classes (signs) included in this study are detailed in the Appendix. In some instances, the hands obscure one another, as seen in signs like "DOCTOR (دکتر)" (doctor), "JOZVEH (جزوه)" (notebook), and "KAM (کم)" (low), while others involve facial expressions, such as "DIDAN (دیدن)" (to see) and "JOGHRAPHIA (جغرافیا)" (geography). Additionally, some signs are compound forms created by combining two consecutive signs, like "OSTAD (استاد)" (teacher). Notably, these constituent signs are also included in our dataset as standalone entries. For instance, the sign for "FAREGH-O-TAHSIL (فارغ التحصیل)" (graduated) is derived from the consecutive signs "DARS (درس)" (lesson) and "TAMAM (تمام)" (end), both of which are represented in the dataset. Furthermore, certain signs involve hand movements in the depth direction, as exemplified by "JOST-O-JOO (جستجو)" (search) and "GOFTAN (گفتن)" (to say). One of the challenges of our dataset is the presence of very similar signs. For example, the signs for "BIMAR (بیمار)" (sick) and "DOCTOR (دکتر)" (doctor) are quite alike, with the only distinction being that "BIMAR" includes an additional gesture of touching the forehead. Additionally, some signs are performed in various ways, further complicating the recognition task.

In Fig. 2, we present examples of various backgrounds from the ISLR101 dataset. The videos showcase a range of lighting conditions, from natural sunlight to artificial illumination, resulting in diverse video frames with varying levels of brightness, shadows, and illuminated areas.

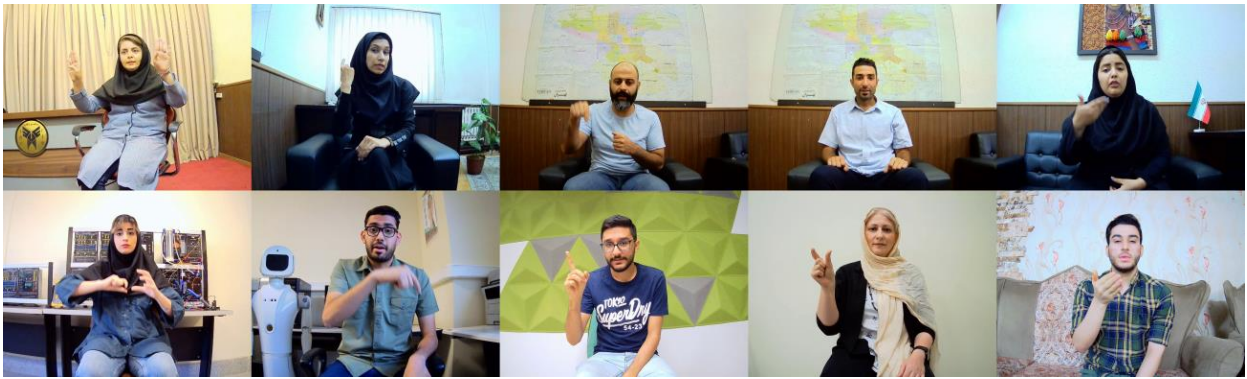

Fig. 2. Example frames from the ISLR101 dataset showcasing different lighting conditions and background variations.

In our dataset, signs are performed by 10 different signers; 3 are Sign language translators, 3 are individuals with hearing problems, and 5 are trained signers who learned the signs in our dataset. Regarding gender, 5 of these signers are male and 5 are female.

Figures 3 and 4 illustrate the distribution of samples across both signs and signers. As depicted, the dataset is balanced in terms of sign distribution. However, the total number of samples recorded for some signers exceeds that of others. The imbalance in the dataset, caused by one signer (signer 6) contributing more videos, introduces a valuable challenge for sign language recognition systems. It tests their ability to generalize across varying signing styles and ensures that models do not overly rely on patterns specific to a single signer, ultimately enhancing their robustness and real-world applicability.

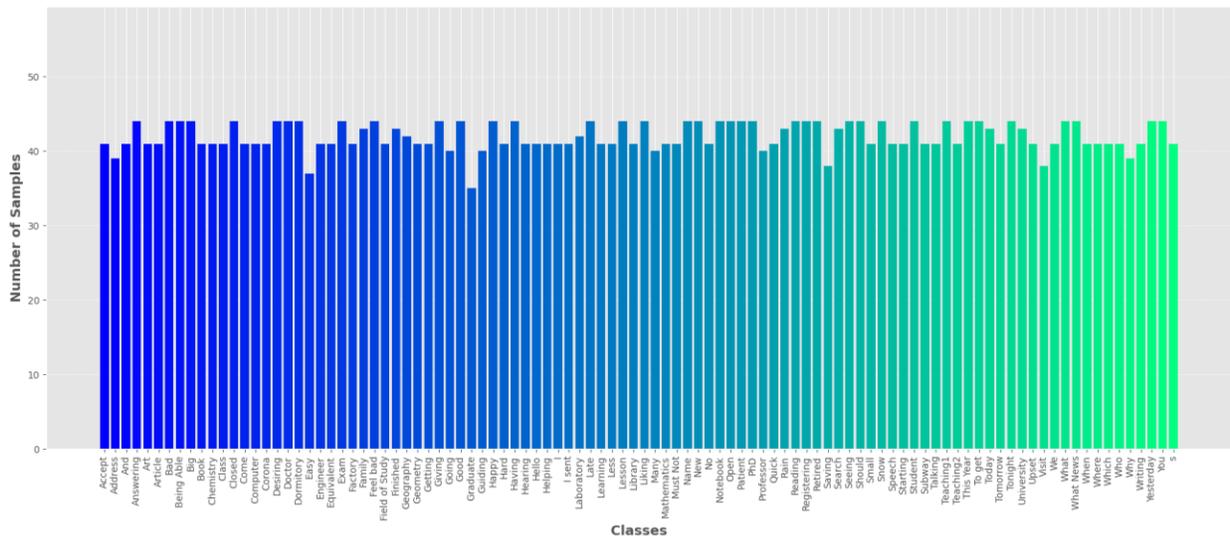

Fig. 3. Number of Video Samples per Class in a 101-Class ISLR101.

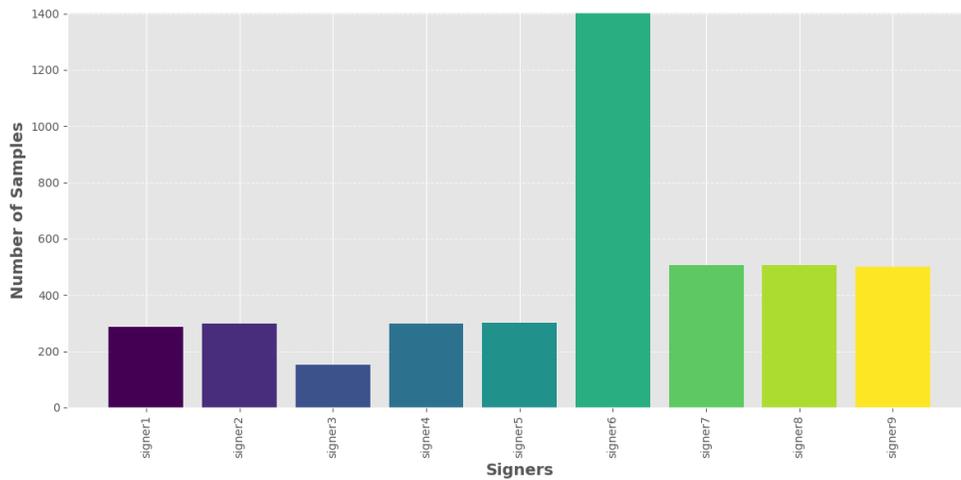

Fig. 4. Number of Video Samples per signer in a 101-Class ISLR101.

## 4. Baseline Models

In this section, we introduce the baseline models, categorized into two groups: visual appearance-based and skeleton-based models. This approach allows us to not only evaluate the applicability of our collected ISLR101 dataset but also to assess the performance of deep learning models across different input modalities for sign recognition.

### 4.1. Visual Appearance-Based Framework

As outlined in Section 2, visual appearance-based frameworks typically consist of three key components: a spatial modeling module, a temporal modeling module, and a classifier. In our baseline model, we design the architecture as follows: 2D CNN + Transformer + Classifier.

For the spatial feature extraction, we adopt MobileNet-V2, which is pretrained on the ImageNet dataset. MobileNet-V2 is chosen due to its balance between low latency and high computational efficiency, making it well-suited for real-time applications. The Transformer is utilized for temporal modeling, as it excels at capturing long-term dependencies, making it effective for tasks requiring sequential understanding of data. The final component of the model is a neural network classifier, implemented as a fully connected layer.

#### 4.1.1 Visual Appearance-Based Framework Data Pre-processing

For the training, we implement stochastic frame dropping, using a dropping ratio of 0.5. During this process, we randomly select 50% of the frames from the original video clip. During testing, an equal 50% of the frames are uniformly chosen from video. This approach effectively reduces computational complexity and eliminates frame redundancy. The videos are resized to 224 × 224. During training, we apply random cropping and random affine (10 degrees) for data augmentation. During testing, we solely utilize center cropping. We split the samples of a gloss into the training, validation, and testing sets following a ratio of 7:1:2.

#### 4.1.2 Implementation details

In the temporal modeling module, we configure the Transformer with a hidden size of 1280, 8 attention heads, and 2 self-attention blocks. We employ a non-trainable positional encoding to effectively capture the temporal order of frames. The classifier is a fully connected network with a single hidden layer consisting of 512 units. ReLU is used as the activation function throughout the network, and cross-entropy is chosen as the loss function. The complete architecture is depicted in Figure 5a.

For training, we utilize 2 NVIDIA A100 GPUs with a batch size of 8. The model is trained using a dropout rate of 0.3, weight decay of $5\times10^{-5}$, and an initial learning rate of $10^{-4}$. The learning rate is reduced by a factor of 0.1 at epochs 15 and 30 to improve convergence. We train all the models with the Adam optimizer [55].

### 4.2. Skeleton-based Frameworks

We utilize the ST-TR model, as outlined in Section 2, and evaluate four frameworks: S-TR, T-TR, ST-TR, and ST-TR-1s (refer to the following subsections of the paper for detailed descriptions of the networks and configurations).

**4.2.1 Spatial Transformer Stream (S-TR)**

The Spatial Transformer Stream (S-TR) applies SSA at the skeleton level to capture spatial relationships between joints. The output from the SSA module is passed through a temporal convolutional network layer (TCN) with a temporal kernel to extract time-relevant features. Additionally, batch normalization and skip connections are employed, with the input summed with the output of the SSA module to enhance the model's performance and stability.

**4.2.2 Temporal Transformer Stream (T-TR)**

The Temporal Transformer Stream (T-TR) focuses on capturing temporal relationships between the frames. Each layer begins with a graph convolution (GCN) to model spatial dependencies, followed by a TSA module. The TSA operates on graphs that link the same joint across all time steps (e.g., all frames of the left foot or right hand) to capture temporal dynamics.

**4.2.3 Two-Stream Spatial Temporal Transformer Network (ST-TR)**

The Two-Stream Spatial Temporal Transformer Network (ST-TR) architecture combines two streams: S-TR for spatial modeling and T-TR for temporal modeling. After separate end-to-end training, the outputs of the two streams are fused by summing their softmax scores to produce the final prediction.

**4.2.4 SSA and TSA in a Single Stream (S-TR-1s)**

We use the integration of SSA and TSA within a single-stream architecture, designated as S-TR-1s. In this configuration, feature extraction employs GCN and TCN modules, with each layer (starting from the 4th) defined as $ST - TR - 1s(x) = TSA(SSA(x))$.

**4.2.5 Skeleton-Based Framework Data Pre-processing**

In this work, we extract 67 keypoints (including 25 body joints and 21 keypoints for each hand) from each frame using OpenPose. For each keypoint, we retrieve its 2D coordinates along with the confidence score indicating the reliability of the detection. We then concatenate all keypoints' 2D coordinates and confidence scores into an array, which serves as the input to our network.

**4.2.6 Implementation details**

The networks are trained on a single NVIDIA A100 GPU. We adhere to the original training configuration, with the exception of using a batch size of 32 and an initial learning rate of $10^{-3}$. The architecture is illustrated in Figure 5b.

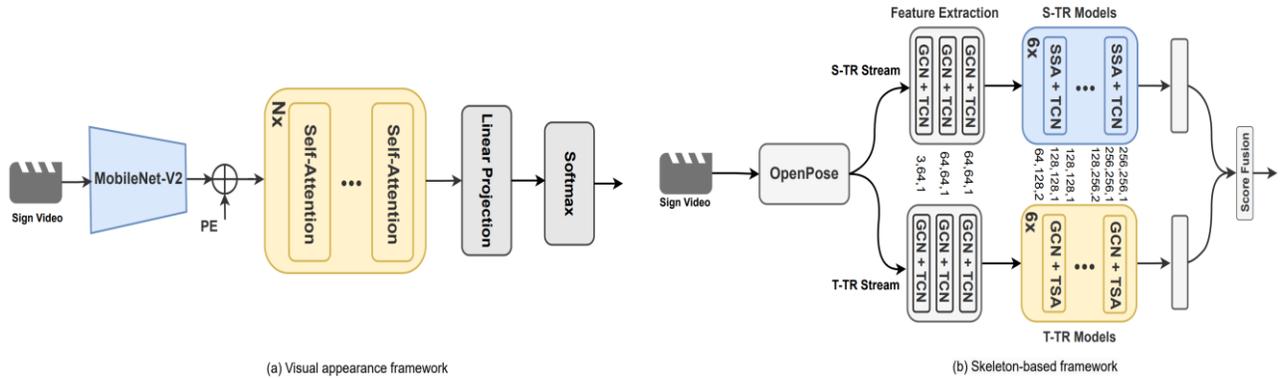

Fig. 5. Illustration of visual appearance-based model and skeleton-based model.

## 5. Results and Discussion

We evaluate the models using the mean top-1 classification accuracy across all sign instances. Table 2 presents the performance of our baseline models based on the pose data and visual appearance. The results show that the visual appearance-based framework achieves higher classification accuracy compared to the pose-based framework. This improvement is expected, as the pose data lacks critical information such as the signer's facial expressions. However, it is important to note that the training process for the pose-based model is nearly 20 times faster.

Table 2. Evaluation results of the baseline models on the ISLR101 dataset.

| Method | No. of Params [$\times 10^6$] | Top-1 Accuracy (%) |
| --- | --- | --- |
| ST-TR | 5.20 | 94.02 |
| ST-TR-1s | 1.98 | 91.58 |
| S-TR | 3.21 | 93.58 |
| T-TR | 1.99 | 85.49 |
| VABF | 28.78 | 97.01 |

The results demonstrate the superior performance of the two-stream architecture, ST-TR, achieving an accuracy of 94.02%, compared to S-TR at 93.58%, ST-TR-1s at 91.58%, and T-TR at 85.49%. This indicates that combining SSA and TSA effectively enhances the model's ability to capture both spatial and temporal relationships. Notably, ST-TR outperforms ST-TR-1s by 2.44%, highlighting the benefits of maintaining separate streams for SSA and TSA.

## 6. Limitations and Future Work

In future work, we plan to enhance the spatial and temporal attention mechanisms in our models to improve robustness in complex environments. Furthermore, we will investigate more advanced learning techniques aimed at better distinguishing visually similar signs, thereby boosting classification accuracy.

## 7. Conclusion

In this paper, we introduce ISLR101, a new large-scale isolated Iranian Sign Language dataset. ISLR101 offers a range of challenges compared to existing sign language datasets. To the best of our knowledge, it is the first publicly available large-scale ISL dataset featuring diverse backgrounds and multiple signers. We provide training, validation, and test set splits to ensure fair model comparisons, along with pose information extracted using OpenPose, all of which are publicly accessible for researchers. We evaluated both visual appearance-based and skeleton-based models on the ISLR101 dataset. The visual appearance-based model, which employs MobileNet-V2 for spatial feature extraction and a Transformer for temporal modeling, outperformed the skeleton-based models, due to the richer data available in the visual modality. Among the skeleton-based approaches, the ST-TR model, which leverages separate streams for Spatial and Temporal Self-Attention, delivered the best performance. These findings highlight the effectiveness of combining spatial and temporal attention for sign language recognition.

**Acknowledgment**

The funding for this study was provided by the "Iranian National Science Foundation (INSF)" (http://en.insf.org/). We would like to express our gratitude to our friend Mr. Ali Ghadami for his cooperation and helps with the data collection, as well as all the signers/participants from the Islamic Azad University, International Fereshtegan Branch, Tehran, Iran.

**Statements & Declarations**

**Conflict of interest**

Author Alireza Taheri has received a research grant from the "Iranian National Science Foundation (INSF)" (http://en.insf.org/) (Grant No. 4031030)". The author Hossein Ranjbar declares that he has no conflict of interest.

**Availability of data and material (data transparency)**

All data from this study (i.e., the ISLR101 dataset) are available in the Social & Cognitive Robotics Laboratory archive.

**Code availability:**

All of the codes are available in the archive of the Social & Cognitive Robotics Laboratory.

**Authors' contributions:**

Both authors contributed to the study's conception and design. Material preparation, data collection, and analysis were performed by Hossein Ranjbar. Alireza Taheri supervised this research. The first draft of the manuscript was written by Hossein Ranjbar; and both authors commented on previous versions of the manuscript. Both authors read and approved the final manuscript.

**Ethical Approval**

Ethical approval for the protocol of this study was provided by the IPM - Institute for Research in Fundamental Sciences.

**Consent to participate:**

Informed consent was obtained from all individual participants included in the study.

**Consent for publication:**

The authors affirm that human research participants provided informed consent for publication of all images. All of the signers/participants have consented to the submission of the results of this study to the journal.

**Appendix**

Address (آدرس)

Came (Amad, آمد)

Easy (AsAn, آسان)

Laboratory (AzmAeshgAh, آزمایشگاه)

PhD (doktorA, دکتری)

Rain (bArAn, باران)

Must (bAyad, باید)

Open (baz, باز)

Visit (bAzdid kardan, بازدید کردن)

Retired (bAzneshasteh, بازنشسته)

Bad (bad, بد)

Dislike (bad Amadan, بد آمدن)

Snow (barf, برف)

Closed (basteh, بسته)

Achieve (bedast Amadan, به دست آمدن)

Patient (bimar, بیمار)

Big (bozorg, بزرگ)

What's up (che khabar, چه خبر)

Why (cherA, چرا)

What is it? (chiyeh, چیه)

Class (kelAs, کلاس)

Computer (kamputer, کامپیوتر)

Corona (korona, کرونا)

Give (dAdan, دادن)

University (dAneshgAh, دانشگاه)

Student (dAneshjoo, دانشجو)

Have (dAshtan, داشتن)

Lesson (dars, درس)

See (didan, دیدن)

Late (dir, دیر)

Yesterday (diroz, دیروز)

Doctor (doktor, دکتر)

Like (dost dAshtan, دوست داشتن)

Today (emroz, امروز)

This year (emsAl, امسال)

Tonight (emshab, امشب)

Exam (emtehAn, امتحان)

Name (esm, اسم)

Tomorrow (fardA, فردا)

Graduate (faregh ol tahsil, فارغ‌التحصیل)

Send (ferstAdam, فرستادم)

Receive (gereftan, گرفتن)

Accepted (ghabol shodan, قبول شدن)

Plural (hA, ها)

Talk (harf zadan, حرف زدن)

Geometry (hendeseh, هندسه)

Art (honar, هنر)

Answer (javAb dAdan, جواب دادن)

Geography (jeography, جغرافیا)

Search (jost o jo, جست‌وجو)

Notebook (jozveh, جزوه)

Workshop (kargAh, کارگاه)

Few (kam, کم)

Book (ketAb, کتاب)

Library (ketAb khAneh, کتاب‌خانه)

When (key, کی)

Dormitory (khAbgAh, خوابگاه)

Read (khAndan, خواندن)

Want (khAstan, خواستن)

Family (khAnevAdeh, خانواده)

Good (khob, خوب)

Happy (khosh hAl, خوشحال)

Who (ki, کی)

Small (kochak, کوچک)

Which (kodAm, کدام)

Where (kojA, کجا)

Help (komak kardAn, کمک کردن)

We (mA, ما)

Article (maghAleh, مقاله)

I (man, من)

Metro (metro, مترو)

Grade (moadel, معدل)

Engineer (mohandes, مهندس)

Sad (nArAhat, ناراحت)

No (na, نه)

Must not (nabAyad, نباید)

Write (neveshtan, نوشتن)

Professor (ostAd, استاد)

Guide (rahnamAi kardan, راهنمایی کردن)

Go (raftan, رفتن)

Field of Study (reshteh tahsili, رشته تحصیلی)

Mathematics (riyAzi, ریاضی)

Register (sabt nAm kardan, ثبت‌نام کردن)

Hard (sakht, سخت)

Hello (salAm, سلام)

Hear (shenidan, شنیدن)

Chemistry (shimi, شیمی)

Start (shoroe shodan, شروع شدن)

Speak (sohbat kardan, صحبت کردن)

Speech (sokhAnrAni, سخنرانی)

New (tAzeh, تازه)

Teach (tadris kardan, تدریس کردن)

Finish (tamAm shodan, تمام شدن)

Can (tavAnestan, توانستن)

You (to, تو)

And (va, و)

Teach (yAd dAdan, یاد داد)